\documentclass[sigconf]{acmart}
\usepackage{fancyhdr}
\renewcommand\footnotetextcopyrightpermission[1]{} 
\usepackage{multirow}
\usepackage{stfloats}
\AtBeginDocument{%
  \providecommand\BibTeX{{%
    \normalfont B\kern-0.5em{\scshape i\kern-0.25em b}\kern-0.8em\TeX}}}

\begin{document}

\title{Simple Contrastive Representation Adversarial Learning for NLP Tasks}

\author{Deshui Miao}
\authornote{Both authors contributed equally to this research.}
\email{3120190789@bit.edu.cn}
\orcid{1234-5678-9012}
\author{Jiaqi Zhang}
\authornotemark[1]
\email{bingfa.zjq@alibaba-inc.com}
\affiliation{%
  \institution{Beijing Institute of Technology}
  \streetaddress{number 5 of zhongguancun sourth street}
  \city{China}
  \state{Beijing}
  \country{China}
}

\author{Wenbo Xie}
\affiliation{%
  \institution{Alibaba Group}
  \city{Beijing}
  \country{China}}
\email{daoqian.xwb@autonavi.com}

\author{Jian Song}
\affiliation{%
  \institution{Alibaba Group}
  \city{Beijing}
  \country{China}}
\email{james.songjian@alibaba-inc.com}

\author{Xin Li}
\affiliation{%
  \institution{Alibaba Group}
  \city{Beijing}
  \country{China}}
\email{beilai.bl@alibaba-inc.com}

\author{Lijuan Jia}
\affiliation{%
  \institution{Beijing Institute of Technology}
  \city{Beijing}
  \country{China}}
\email{jlj@bit.edu.cn}

\author{Ning Guo}
\affiliation{%
  \institution{Alibaba Group}
  \city{Beijing}
  \country{China}}
\email{ning.guo@alibaba-inc.com}

\renewcommand{\shortauthors}{Deshui Miao and Jiaqi Zhang, et al.}

\begin{abstract}
  Self-supervised learning approach like contrastive learning is attached great attention in natural language processing.
  It uses pairs of training data augmentations to build a classification task for an encoder with well representation ability.
  However, the construction of learning pairs over contrastive learning is much harder in NLP tasks.
  Previous works generate word-level changes to form pairs, but small transforms may cause notable 
  changes in the meaning of sentences as the discrete and sparse nature of natural language.
  In this paper, adversarial training is performed to generate challenging and harder learning adversarial examples over the embedding space of NLP as learning pairs.
  Using contrastive learning improves the generalization ability of adversarial training because contrastive loss can uniform the sample distribution.
  And at the same time, adversarial training also enhances the robustness of contrastive learning.
  Two novel frameworks, supervised  contrastive adversarial learning (SCAL) and unsupervised SCAL (USCAL), are proposed, which yield learning pairs by utilizing the adversarial training for contrastive learning.
  The label-based loss of supervised tasks is exploited to generate adversarial examples while unsupervised tasks bring contrastive loss.
  To validate the effectiveness of the proposed framework, we employ it to Transformer-based models for natural
  language understanding, sentence semantic textual similarity, and adversarial learning tasks.
  Experimental results on GLUE benchmark tasks show that our fine-tuned supervised method outperforms BERT$_{base}$ over 1.75\%.
  We also evaluate our unsupervised method on semantic textual similarity (STS) tasks, and our method gets 77.29\% with BERT$_{base}$.
  The robustness of our approach conducts state-of-the-art results under multiple adversarial datasets on NLI tasks.
\end{abstract}




\keywords{self-supervised learning, contrastive learning, adversarial training, natural language}

\maketitle
\begin{figure*}[h]
  \centering
  \includegraphics[width=\textwidth,height=0.3\textwidth]{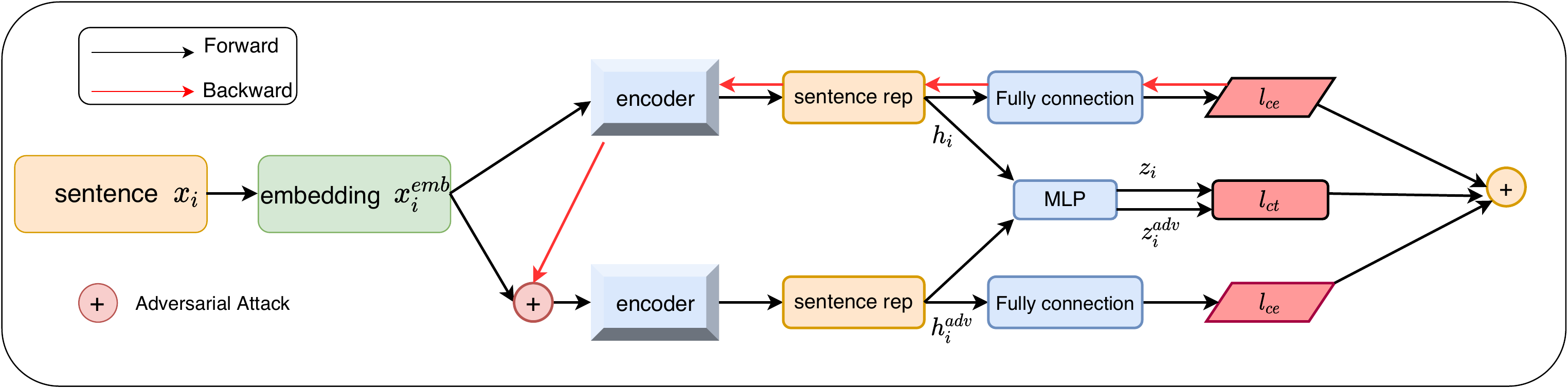}
  
  \caption{Supervised Contrastive Adversarial Learning framework (SCAL). SCAL both minimizes supervised classification loss and minimizes a similarity loss between $z_{i}$ and $z_{i}^{adv}$. $z_{i}^{adv}$ is
  the representation of adversarial embeddings, which is generated by calculating of the gradient of $x_{i}^{emb}$ and then adding to $x_{i}^{emb}$. In the representation space, Contrastive 
  Learning push together the positive pairs and pull away negetive pairs. Both encoders and fully connection layers in all branches are sharing the weight, 
  and are drawn into two structures for good understanding. At the end of training, $h_{i}$ is used as the sentence representation.}
\end{figure*}
\section{Introduction}
Pre-trained Language Models (PLMs) \cite{xlnet} such as BERT \cite{bert} and RoBERTa \cite{roberta} have shown significant impact on various Natural Language Processing (NLP) tasks such as text classification \cite{finetunebert}, 
sentence representation \cite{bertsentence}, and machine translation \cite{machinetranslation}.
By designing effective self-supervising learning objectives, such as masked language modeling like BERT, PLMs are able to catch semantic features in text,  language representations for various downstream NLP tasks.
Specifically, PLMs do not directly produce useful sentence representations for downstream tasks while fine-tuning PLMs with labeled data in downstream training set is a common step to construct better representations.

Recently, self-supervised learning contrastive learning has drawn much attention and many excellent works have been proposed \cite{moco, simclr, mocov2,siamese,byol}.
Contrastive learning (CL) is a paradigm which learns an effective feature representation for discriminative models by positive pairs and negative pairs.
The model is trained with a contrastive loss by comparing positive pairs and negative pairs which are generated by original input using various transformation methods.
This leads the embedding representation generated by pairs from the same instances to push together while from different instances to pull away.
How to get the positive pairs has caught many researchers' great attention like SimCLR \cite{simclr}, which compares different combinations of image transformation methods such as rotation, scaling and, random cropping.
Those data augmentation methods have achieved great success for CL in compute vision (CV) tasks because CL benefits more from constructing effective positive pairs for invariant instance learning. 
Due to the discrete and sparse nature of natural language, it is difficult to generate label-preserving augmentations of the same instance for contrastive learning in NLP tasks.
Many recent works proposed different data transformation ways for contrastive learning.
\cite{DCLT} proposed a new framework using contrastive learning to learn meaningful sentence representations with unlabeled data, which offers a great competitive performance to supervised methods on some tasks.
But the method needs to build negative pairs carefully and how to use CL efficiently is still has great room for improvement.
And BYL was proposed by \cite{byl} which is a framework like BYOL in compute vision (CV), which only uses the augmentations of different examples to negative pairs in the same batch.
The augmentation method of BYL is back-translation but this method can easily change the meaning of original sentences.
SimCSE \cite{simcse} uses simple dropout of PLMs like BERT to construct different embeddings as positive pairs of original examples.
Sometimes in NLP tasks dropout and other augmentation methods like random delete may change the information of original short sentences, so how to generate positive pairs is now a hotspot.
In this work, our goal is to take a further step towards searching an efficacious data augmentation medium through adversarial training.

Adversarial Training is introduced in \cite{adtraining} and is usually considered as an efficacious way to against attacks.
It is performed by generating confusing examples to perturb the model and maximize the total loss on the target model, which can improve the robustness of the network greatly \cite{Intriguing}.
The confusing examples are usually generated by adding a small gradient-based perturbation to the original data \cite{fgsm}.
Various approaches are proposed to perturb target models over the past few years such as Fast Gradient Sign Method (FGSM) \cite{fgsm} which applies a perturbation in the gradient direction with a small parameter, 
Projected Gradient Descent (PGD) \cite{pgb} which maximizes the loss over $k$ iterations and so on.
But these methods are not suitable to apply directly to NLP tasks because the sentence is word-based and the gradient can not add to it directly.
\cite{embedad} proposed a way that adds the perturbation to word embeddings of the encoder on text classification tasks, which is gradually a common way to perform adversarial training in NLP tasks. 
Although adversarial training has been widely used in many tasks, it may hurt generalization \cite{alum, hurtgeneral}.
Recent works focus more on the impact of generalization of adversarial training like FreeLB \cite{freelb}, and we use the feature of contrastive learning to enhance the generalization of adversarial training in our work.

Contrastive learning with adversarial training is learned in many computer vision papers \cite{adcl1, adcl2, byorl, adcl3}.
\cite{adcl2} addresses to define a new adversarial training for self-supervised learning while \cite{adcl1, byorl, adcl3} aim to obtain a more robust model.
There are also some works about contrastive and adversarial learning in NLP \cite{adco, adclfortextgeneration, calplms, cline, adclNLP}.
The way to use adversarial training is quite different such as \cite{adco} proposes a method to create adversarial examples on word-level 
and \cite{cline} uses external semantic knowledge to generate negative instances.
\cite{adclNLP} performs adversarial training with a perturbation on  one token embedding while our works add perturbations to all input embeddings.
The above all contrastive adversarial training methods only focus on the supervised tasks, and we care about both supervised and unsupervised circumstances.
Our contrastive adversarial training is different from other works in: (i) For different tasks, two task-based frameworks are proposed,.
(ii) Adversarial training in proposed frameworks is applied to the embedding-level. (iii) Different loss functions are used to generate adversarial examples.

In this work, we seek a way to generate meaningful positive pairs and challenging negative pairs.
This is because harder positive pairs can lead to the largest optimization cost and make the whole model get better embeddings.
We aim to get these hard examples by leveraging adversarial training, which are concertized to attack models and thus can be regarded as a sufficient way to generate the most challenging examples.
In the meanwhile, adversarial training can also enhance the robustness of the model.
Contrastive learning can be used to improve the performance of supervised tasks and unlabeled tasks, which has been proven in recent studies. 
Therefore, a SCAL framework used PLMs is developed for. for fine-tuning on NLU tasks like GLUE.
In addition, in order to overcome the collapse of PLMs like BERT, an unsupervised contrastive adversarial learning framework is built to obtain better sentence representations, which has proven more useful than existing methods.
In the supervised framework, it is easy to generate adversarial examples by cross entropy loss while it is difficult to execute. .
We find that it is possible to treat the contrastive learning as an instance classification task, therefore, sensitive attacks can be generated by contrastive losses in an unsupervised framework.  
The introduction of contrastive learning improves the data separability and model generalization performance of adversarial learning. At the same time, the adversarial training 
strategy also assists contrastive learning to increase the learning difficulty and robustness of the model.
Overall, our method in supervised tasks can be treated as an auxiliary means and in unsupervised tasks is a framework way.

\section{Related work}

The main component of our work is contrastive adversarial learning, so the survey of previous work mainly focuses on these two parts.

\subsection{Contrastive Learning}
Recently, contrastive learning has been widely used in self-supervised learning, which learns an encoder to represent each image in the training set.
And a good feature representation should have the ability to identify the 
same objects while distinguishing itself from others \cite{ctrloss}.
Contrastive learning is first introduced in computer vision (CV) and many papers employ image transformations such as rotation, color change and cropping to generate two augmentation version of original image as positive pair, 
which are close to each other in the representation space \cite{ctrloss, wu2018unsupervised, simclr, moco, mocov2, byol, siamese, sermanet2017time}.
Going beyond unsupervised contrastive learning, it can also be applied in supervised tasks to leverage the labeled training set sufficiently \cite{supervisedctr}.
SimCLR \cite{simclr} compares different combinations of image augmentation method and uses a much simpler way to train the model, but it uses bigger batch-size and epoch because the model needs to see more 
examples which comsume a lot of computing resources.
MOCO \cite{moco, mocov2} uses a dynamic queen to store a lot  examples where the model push the latest embeddings to a memory bank and pop the oldest data.
Ways using a dynamic queen and big batch-size is a common way to let model see more examples before BYOL which just train the model by discriminating instance within a minibatch.

This learning paradigm has been successfully employed to numerous NLP tasks \cite{supervisedplm}.
The key component of contrastive learning is how to generate the positive pair while generating the positive pair in NLP tasks is quite difficult.
Traditional data augmentations such as back translation, word and span deletion and sequence cropping were used frequently while applying contrastive learning \cite{byl, adco, cline}.
SimCSE \cite{simcse} proposed a simple way, dropout in PLMs, to improve the ability of sentence representation of PLMs.
Our work also use dropout to generate different views of one instance but the adversarial examples are also added as positive pairs to contrastive loss.
This not only increases the difficulty of training, but also makes the model more robust and expressive.

\subsection{Adversarial Training and Adversarial Attack}
Adversarial training means that the network in trained with clean and adversarial examples to defend against attacks and improve the robustness of network, 
which has been applied in many supervised scenarios such as object detection \cite{adob}, segmentation \cite{adseg} and image classification \cite{adclassification}.
Based on the setting of adversarial training, to complete the training, adversarial examples must be generated by the network and the clean examples, which make the network predict the wrong class label \cite{adexamples, fgsm, adexample2}.
Researchers have proposed a variaty of adversarial training methods and adversarial attack ways.
Word-level substitution \cite{white, nlpadversarial, wordattack} and sentence-level \cite{PAWS} rephrasing are the typical textual adversarial attacks to perform semantic conservation that fools the model.
For instance, \cite{fgsm} proposed the Fast Gradient Sign Method (FGSM), which generates a perturbation added to target sample, to increase the loss, and use the generations and the original sample to train the model.
Later gradient-based attack works proposed iterative ways to attack the examples with improved framework \cite{FGM,pgb, freelb, smart}.
Follow-up works \cite{adtraining} begin to use adversarial training, a way familied in image tasks, to improve the network robustness.
In our work, we use the FGSM \cite{fgsm} and Fast Gradient Method (FGM) \cite{FGM} to generate adversarial attacks and then add it to word embeddings of original instance for constructing positive pairs.
There are many perfect works in computer vision to use contrastive learning and adversarial training \cite{adcl1, adcl2, byorl, adcl3}.
And they are also used in various NLP tasks \cite{adco, adclfortextgeneration, calplms, cline}.
Our work focus on the adversarial attack to embedding-level of the model encoder, and regard the adversarial examples as positive pairs of the original instance.
Two frameworks generating adversarial examples are proposed under different tasks.
The proposed models achieve great improvement both on performance and robustness.
\begin{figure*}[ht]
  \centering
  \includegraphics[width=\textwidth,height=0.3\textwidth]{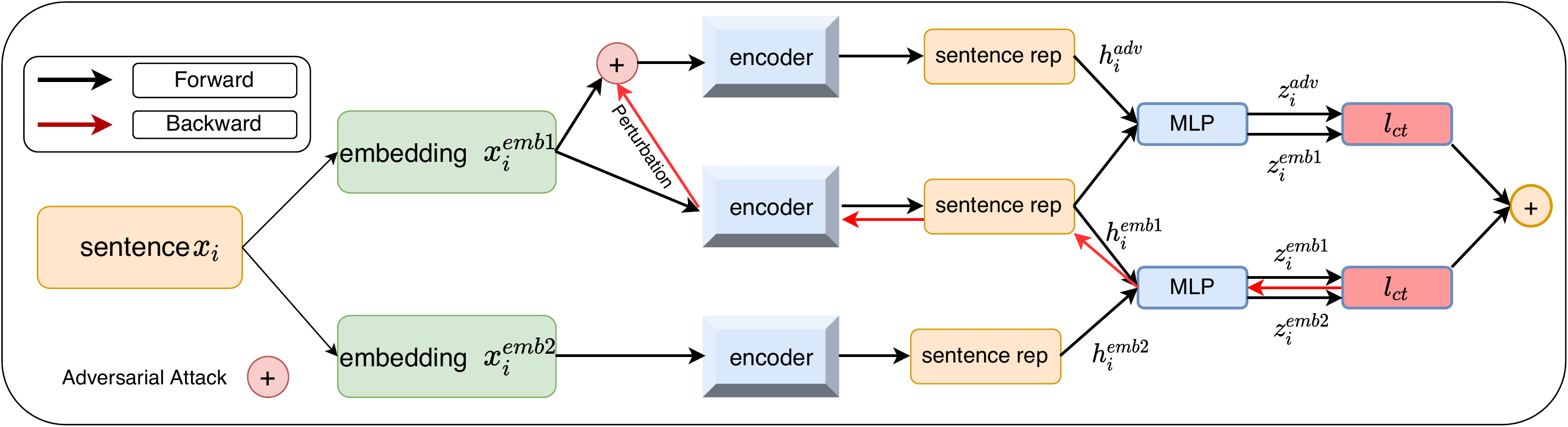}
  \caption{Unsupervised Contrastive Adversarial Learning framework. $x_{i}^{emb1}$ is the embedding of the original data $x_{i}$ and $x_{i}^{emb2}$ is the embedding of the augmentation of $x_{i}$. USCAL minimizes two combined similarity loss for $z_{i}^{emb1}$, $z_{i}^{emb2}$ and $z_{i}^{adv}$.  
  $z_{i}^{adv}$ is the representation of adversarial embeddings, which is generated by calculating of the gradient of contrastive loss between $z_{i}^{emb1}$ and $z_{i}^{emb2}$ with respect to  $x_{i}^{emb1}$.
  In the representation space, Contrastive Learning push together the positive pairs pull away negetive pairs. 
  Same as SCAL, note that, whenever more than one encoder and MLP branches co-exist in one framework, they by default share all weights. 
  At the end of training, $h_{i}^{emb1}$ is used as the sentence representation.}
\end{figure*}

\section{Methods}
We now describe how two frameworks are established to improve the robustness and performance of PLMs, using FGSM attack methods generated by different losses in two frameworks.
Before describing framework, we first briefly introduce the adversarial training and self-supervised contrastive learning.
\subsection{Adversarial Training And Adversarial Attack Methods}
We first give the definition of tasks with adversarial attacks under supervised settings.
Consider a training set $\mathbb{D} = \{ \textit{X}, \textit{Y} \} $, where $x\in \it X$ is the training example and $y\in \it Y$ is the relevant label, and a supervised model with a deep representation network is defined as
network $f_{\theta}: \textit{X} \rightarrow \textit{Y}$, where $\theta $ is the learnable parameter of the model.
With the set of training dataset and learning models, adversarial attacks attempt to search a confusing examples using perturbation to fool the target model, which maximizes the total loss.
Based on the supervised settings, we use the cross entropy loss to demonstrate the expression of  adversarial attacks with as follows:
\begin{equation}
  \delta ^{*} = arg \max_{\delta }  \textit{L}_{ce}(x+\delta, y; \theta ) \quad s.t. \Vert  \delta \Vert _{p} < \varepsilon,
\end{equation}
where $\delta $ is an adversarial perturbation generated by $\textit{L}_{p}$ norm and smaller than $\varepsilon $.
There are a pair of gradient-based algorithms proposed to solve this optimization and our work focuses on FGSM and FGM.
FGSM leverages sign function to control the direction for searching adversarial examples which are generated by:
\begin{equation}
  x ^{adv} = x + \delta  = x + \varepsilon  * sign(\bigtriangledown_{x} \textit{L}_{ce}(x, y; \theta )) \quad s.t. \Vert  \delta \Vert _{p} < \varepsilon,
\end{equation}
where $x^{adv}$ is the adversarial example using FGSM and $\varepsilon $ is the positive learning rate for generate adversarial examples $x^{adv}$.
FGM is proposed with different gradient processing where it is scaled according to specific gradients to get better adversarial samples.
\begin{equation}
  x ^{adv} = x + \delta  = x + \varepsilon  * \frac{\bigtriangledown_{x} \textit{L}_{ce}(x, y; \theta )}{\Vert \bigtriangledown_{x} \textit{L}_{ce}(x, y; \theta ) \Vert_{2} }  
  \quad s.t. \Vert  \delta \Vert _{p} < \varepsilon,
\end{equation}
These two algorithms are leveraged both in our supervised model and unsupervised model.

\subsection{Self-supervised Contrastive Learning}
In the setting of self-supervised learning, the training data is unlabeled, like $\mathbb{D} = \{ \textit{X} \}$, each instance $x$ in dataset is mapped into two different augmentation views 
as positive pair $(x^{p}, x^{q})$.
An excellent model needs to push examples from the same positive pair closer while farther away the  pairs in representation space.
Contrastive learning seeks to learn an effective encoder to represent original data by minimizing the contrastive loss, maximizing the similarity, which is  defined in InfoNCE as:
\begin{equation}
  \mathcal{L}_{ct} = -\log \frac{\exp(sim(x_{i}^{p} x_{i}^{q}/\tau)) }{\sum_{k=1}^{N}  \exp(sim(x_{i}^{p} x_{k}^{q}/\tau)) } , p,q \in  \chi 
\end{equation}
where $x_{i}^{p}$, $x_{i}^{q}$ are the embeddings of different augmentations from the same instance $x_{i}$, which are considered as the positive pairs; and $\tau $ is the positive temperature constant.
$sim()$ calculates cosine similarity between two vectors.
$N$ is the batch-size and $\chi $ is a set of transformation methods.
The choice of $p, q$ from $\chi $ is confirmed critical in \cite{simclr} for contrastive learning.
In our works, we use different transformations between supervised method and unsupervised framework.
The adversarial example is considered to be the positive pair of original sample in supervised model while first dropout is leveraged in unsupervised work to generate positive pairs 
and then adversarial examples also be used as positive pair.
Through the expression of (4), the key problem is the differents between two augmentations, so we use adversarial attack generates difficult positive pairs and training the model adversarially.

\begin{table*}[ht]
  \caption{GLUE tasks}
  \centering
  \begin{tabular}{llllll}
    \toprule
    Corpus    &  Task     &  Train    &  Domain    &  Classes    &  Metric \\
    \midrule
    \multicolumn{6}{c}{Similarity and Paraphrase Tasks }\\
    \midrule
    MRPC      & paraphrase     & 3.7k    & news   & 2    & F1-score   \\
    QQP      & paraphrase     & 364k    & online QA   & 2    & Accuracy   \\
    \midrule
    \multicolumn{6}{c}{Inference Tasks }\\
    \midrule
    RTE    & textual entailment     & 2.5k    & news/Wikipedia    & 2    & Accuracy      \\
    QNLI    & question answering/textual entailment     & 105k    & Wikipedia    & 2    & Accuracy   \\
    MNLI    & textual entailment    & 393k    & multi-domain  & 3    & Accuracy   \\
    \midrule
    \multicolumn{6}{c}{Single-Sentence tasks }\\
    \midrule
    SST-2   & sentiment analysis    & 67k    & movie reviews   & 2    & Accuracy   \\
    COLA  & grammatical correctness    & 8.5k    & linguistic publications   & 2    & Mcc  \\
    \bottomrule
  \end{tabular}
\end{table*}

\subsection{Supervised Contrastive Adversarial Learning}
How to use contrastive adversarial learning to improve the performance of supervised tasks is the focus of this part.
We introduce the first new framework proposed in supervised tasks where the positive pairs contain clean examples and adversarial examples.
The positive pairs and the negetive pairs are produced by different augmentation methods, but in NLP tasks, augmentation methods usually map in word-level.
The pairs are generated through adversarial attack in word-embedding level in the proposed method.
Adding the adversarial attack to contrastive framework is the vital way to fine-tune the PLMs in downstream tasks.
In our work, we utilize the online work in BYOL to construct our contrastive learning framework and the main structure is shown in Fig. 1.

Given a supervised task with training dataset $\mathbb{D} = \{ \textit{X}, \textit{Y} \} $.
The input sentence $x_{i}$ is first sent to encoder and generate a feature representation $h_{i}$ of $x_{i}$.
The whole network is performed by minimizing the cross entropy loss:
\begin{equation}
  \mathcal{L}_{ce} = -\frac{1}{N}  \sum_{i = 1}^{N}  \sum_{c=0}^{C}   y_{i,c}\log(p(y_{i,c}\vert h_{i}))
\end{equation}
where $y_{i}$ is the label of $x_{i}$ and $C$ is the number of classes.
Then the adversarial perturbation is generated and added to word-embeddings in every examples in batch:
\begin{equation}
  x_{i} ^{adv} = x_{i}^{emb} + \delta  = x_{i}^{emb} + \varepsilon  * \frac{\bigtriangledown_{x_{i}^{emb}} \textit{L}_{ce}(x_{i}, y_{i})}{\Vert \bigtriangledown_{x_{i}^{emb}} \textit{L}_{ce}(x_{i}, y_{i} \Vert_{2} }  
\end{equation}
where $x_{i}^{emb}$ is the embedded example of $x_{i}$.
This is the expression of FGM, obviously FGSM can also be used.
Adversarial examples also go through encoder to produce feature expressions $h_{i}^{adv}$, which textures positive pair with $h_{i}$ and assembles  pairs with other instances in batch.
Like BYOL, $h$ and $h^{adv}$ do not compare directly as they pass a pooler layer ,generating $z$ and $z^{adv}$, to contrastive loss.
\begin{equation}
  \mathcal{L}_{ct} = -\log \frac{\exp(sim(z_{i} z_{i}^{adv}/\tau)) }{\sum_{k=1}^{N}  \exp(sim(z_{i} z_{k})/\tau) } 
\end{equation}
Overall, we describe how to train the encoder which can be different PLMs in supervised tasks.
And the total \textit{Supervised Contrastive Adversarial Learning (SCAL)} objective is formulated by a weighted optimization:
\begin{equation}
  \mathcal{L}_{total} = \frac{1}{2} (\mathcal{L}_{ce}(x,y) + \mathcal{L}_{ce}(x^{adv}, y)) + \alpha * \mathcal{L}_{ct}(x, x^{adv})
\end{equation}
where $\alpha $ is a constant, less than one, that controls the  proportion of contrastive loss.

\subsection{Unsupervised Contrastive Adversarial Learning}
Then the USCAL is considered, which is mostly used to sentence representation.
SimCSE proposed a simple way to use contrastive learning.
It only leverages the dropout masks in Transformers but gets greatly improvement in sentence representation.
In this section, we apply the SimCSE formulation of positive and negetive pairs for the first step and the key contribution of our work is in the second step, where 
the adversarial example is produced by contrastive loss in the first step.
According to BYOL, we can treat the contrastive learning as the prediction tasks, where one augmentation is the label of the other.
So we establish the unsupervised paradigm using contrastive learning and adversarial training.

Similar to SCAL,, the training dataset is set as $\mathbb{D} = \{ \textit{X} \}$.
The input data $x_{i}$ gets two different embeddings $x_{i}^{emb1}, x_{i}^{emb2}$ due to dropout masks, and then go through encoder to get feature representation $h_{i}^{emb1}, h_{i}^{emb2}$ which is 
regarded as positive pair.
Then they are also mapped by a pooler layer sent to contrastive loss.
The different between supervised model and unsupervised model is the generation of adversarial examples as:
\begin{equation}
  x_{i} ^{adv} = x_{i}^{emb1} + \delta  = x_{i}^{emb1} + \varepsilon  * \frac{\bigtriangledown_{x} \textit{L}_{cl}(x^{emb1},x^{emb2})}{\Vert \bigtriangledown_{x} \textit{L}_{cl}(x^{emb1},x^{emb2} \Vert_{2} }  
\end{equation}
which is also the FGM one.
And the total \textit{Unsupervised Contrastive Adversarial Learning} objective is also calculated by a weighted optimization:
\begin{equation}
  \mathcal{L}_{total} = \mathcal{L}_{ct}(x^{emb1},x^{emb2})  + \alpha * \mathcal{L}_{ct}(x^{emb1}, x^{adv})
\end{equation}

\section{Experiments}
In this section, we assess the performance of our proposed SCAL and USCAL framework on different tasks using large pretrained language models like Bert trained on large datasets.
\begin{table*}
  \caption{Short-text classification tasks}
  \centering
  \begin{tabular}{lllllll}
    \toprule
    Dataset     &  Type    &  Classes   &  Average lengths   &  Train  &  test  &  Metric \\
    AG's News     & Topic     & 4   & 44   & 120k  & 7.6k   & Accuracy   \\
    TREC     & Question    & 6   & 11   & 5452  & 500   & Error Rate   \\
    \bottomrule
  \end{tabular}
\end{table*}
\subsection{Training Details}
\begin{table*}[ht]
  \caption{GLUE Results}
  \centering
  \begin{tabular}{llllllllc}
    \toprule
    Model    &  MRPC    &  COLA   &  RTE  &  SST-2 &  QNLI &  QQP  &MNLI(m/mm)&  Avg.\\
    \midrule
    Bert$_{base}$     & 88.6    & 56.3   & 67.5   & 92.0  & 90.1   & 90.7 &83.91/84.10  & 81.65  \\
    Bert$_{base}$+SCAL    & \textbf{92.0}    & \textbf{61.7}   & \textbf{69.7}   & \textbf{92.8}  & \textbf{90.9}  & \textbf{91.4} &\textbf{84.1/84.6} & \textbf{83.4}\\
    \bottomrule
  \end{tabular}
\end{table*}

For a fair comparison with existing works, we adopt Bert and RoBERTa as the main encoder for training.
A pooler layer and a classification layer are added on the top of [CLS] representations for contrastive loss and classification in supervised tasks.
USCAL only adds a pooler layer.
We employ AdamW optimizer with $0.01$ weight decay and warm up learning rate scheduler.
For SCAL, the max sequence length for tokenizer is set to 128 and the batch-size is 32 with BERT and 16 with RoBERTa.
We finetune models for 15 epochs with early stopping strategy and the learning rate for starting is 3e-5.
The temperature constant for contrastive loss is set as 0.05 and the perturbation parameter $\varepsilon \in \{0.1, 0.2, 0.3, 0.4, 0.5\}$.
For USCAL, the batch-size is 64 and the token length is 32, we evaluate our models every 250 training steps on development set of STS-B and save the best checkpoint for the final test.

\subsection{Datasets}
First, the experiments of SCAL is performed on six tasks of GLUE benchmark, including five natural language understanding tasks, question answering/entailment (QNLI), paraphrase (MRPC),
question paraphrase (QQP), textual entailment (MNLI, RTE), and grammatical correctness (CoLA).
Tabel 1 summarizes the summary of these GLUE tasks.

Besides, we test the performance of our model on short-text classification tasks, including AG's News and TREC.
AG's News is a dataset for topic classification with four types of news article: World, Business and Science, Sports
 and Technology.
TREC dataset is for question classification, including open-domain, fact-based questions divided into broad semantic categories.
The detail of these two datasets is shown in Table2.

SentEval contains a pair of STS datasets including the STS task from 2012 to 2016, the STS benchmark (STS-B) and the SICK-Relatedness dataset.
These datasets evaluate the similarity of sentence pairs by scores between 0 and 5, where a higher score indicates higher similarity between two sentences.
Spearman's rank correlation between consine similarity of sentence pairs and the true similarity is used as the evaluation score of our work.
We train the unsupervised model on wiki dataset and then test it on SentEval, following SimCSE.
The best model is saved by evaluating the model on STS-B dataset.

Furthermore, robustness is proved by the dataset of ANLI, used in previous work like InfoBert.
We also follow the training step of InfoBert for a fair comparison.
ANLI is a dataset providing adversarial traing data, which is a large-scale NLI benchmark and collected via an iterative, daversarial, human-and-model-in-the-loop process 
to attack BERT and RoBERTa.
The reason we use it is the ANLI dataset is a powerful and adversarial dataset that easily reduces the accuracy of BERTLarge to 0\%.

\subsection{Models}
\textbf{Baseline}: Two typical model zoos, Bert \cite{bert} and RoBERTa \cite{roberta}, are used as encoder in our model, which contains Bert$_{base}$,Bert$_{large}$, RoBERTa$_{base}$ 
and RoBERTa$_{large}$.

\textbf{Model To Compare} We compare our models with different methods based on the above baseline.

$\diamondsuit$ \textit{SimCSE} \cite{simcse} uses simple dropout to construct positive and  pairs and shows great improvement in sentence representation tasks.

$\diamondsuit$ \textit{FreeLB} \cite{freelb} adopts adversarial training to implement smooth outputs and improve generalization.

$\diamondsuit$\textit{SMART} \cite{smart} performs start-of-the-art generalization and robust in many NLP tasks by fine-tuning PLMs with adversarial and smoothness constraints.

$\diamondsuit$ \textit{ALUM} \cite{alum} conducts adversarial training both in pre-training and fine-tuning stages, which realizes great improvement on a variaty of NLP tasks.

$\diamondsuit$ \textit{InfoBERT} \cite{infobert} applies two mutual-information-based regularizers, information bottleneck regularizer and anchored feature regularizer, 
to train the model, which brings about state-of-the-art robust accuracy over a wide range of adversarial datasets.

Due to the high computational cost for training the whole models, we compare our model to those models with the best result in the original papers.

\begin{table}[ht]
  \caption {ablation Results}
  \centering
  \begin{tabular}{llllllll}
    \toprule
    Model    &  MRPC    &  COLA   &  RTE\\
    \midrule
    Bert$_{base}$     & 88.6    & 56.3   & 67.5 \\
    Bert$_{base}$+CL     & 90.7    & 60.4   & 69.0 \\
    Bert$_{base}$+AT     & 91.0   & 59.0  & \textbf{70.0} \\
    Bert$_{base}$+SCAL    & \textbf{92.0}    & \textbf{61.7}   & 69.7 \\
    \bottomrule
  \end{tabular}
\end{table}

\begin{table}[ht]
  \caption{Results on short-text classification}
  \centering
  \begin{tabular}{ccccccc}
    \toprule
    Model    &  AG-News    &  TREC(Error Rate)  \\
    \midrule
    Bert$_{base}$     & 94.2    & 3.0   \\
    Bert$_{base}$+CL     & 94.7    & 2.0    \\
    Bert$_{base}$+AT     & 95.0   & 2.2   \\
    	
    Bert-ITPT-FiT        &95.2    &3.2\\
    Bert$_{base}$+SCAL    & \textbf{95.3}    & \textbf{1.8}   \\
    \bottomrule
  \end{tabular}
\end{table}

\begin{table*}[ht]
  \caption{Sentence representation predormance on STS tasks, using Spearman's correlation. We have bolded the improved indicators compared to previous works.
  Where $\clubsuit $ indicates the result from \cite{simcse}, $\diamondsuit $ directs \cite{isbert}, $\heartsuit$ from \cite{simcse} and $\spadesuit$ is \cite{consert} }
  \centering
  \begin{tabular}{cccllllll}
    \toprule
    Model                                &  STS12  &  STS13  &  STS14  &  STS15 &  STS16 &  STS-B  &  SICK-R  &  Avg.\\
    \midrule
    Bert$_{base}$$\clubsuit$             & 39.70   & 59.38   & 49.67  & 66.03  & 66.19   & 53.87   & 62.06    & 56.70\\
    Bert$_{base}$-flow$\diamondsuit$     & 58.40   & 67.10   & 60.85  & 75.16  & 71.22   & 68.66   & 64.47    & 66.55\\
    Bert$_{base}$-whitening$\diamondsuit$& 57.83   & 66.90   & 60.90  & 75.08  & 71.31   & 68.24   & 63.73    & 66.28\\
    IS-Bert$_{base}$$\heartsuit$         & 56.77   & 69.24   & 61.21  & 75.23  & 70.16   & 69.21   & 64.25    & 66.58\\
    ConSERT$_{base}$ $\spadesuit$        & 64.64   & 78.49   & 69.07  & 79.72  & 75.95   & 73.97   & 67.31    & 72.74\\
    SimCSE-Bert$_{base}$$\diamondsuit$   & 67.53   & 82.23   & 72.49  & 81.40  & 77.59   & 76.64   & 70.88    & 75.54\\
    Bert$_{base}$+USCAL                   & \textbf{70.61}    & \textbf{82.73}   & \textbf{76.21}   & \textbf{82.61}  & \textbf{77.85}  & \textbf{78.56}  & \textbf{72.48}  & \textbf{77.29}\\
    \hline
    RoBERTa$_{base}$ (first-last avg.)$\diamondsuit$ &40.88 &58.74 &49.07 &65.63 &61.48 &58.55 &61.63 &56.57\\
    RoBERTa$_{base}$-whitening$\diamondsuit$ &46.99 &63.24 &57.23 &71.36 &68.99 &61.36 &62.91 &61.73\\
    SimCSE-RoBERTa$_{base}$$\diamondsuit$ &68.68 &82.62 &73.56 &81.49 &80.82 &80.48 &67.87 &76.50\\
    RoBERTa$_{base}$+USCAL & 68.63 &\textbf{82.95} &\textbf{75.11} &\textbf{82.40} &\textbf{81.59} &\textbf{80.54}     &\textbf{70.11} &\textbf{77.33}\\ 
    \bottomrule
  \end{tabular}
\end{table*}

\subsection{Experiment Result on SCAL}
\textbf{Evaluation on GLUE benchmark} On Glue tasks, we fine-tune our model with encoder BERT$_{base}$ which are trained in large datasets.
Classification loss, Cross Entropy, is applied to both clean and adversarial examples respectively while contrastive loss is used to push together the clean examples and adversarial examples.
We compare our results with the fine-tuned output of original BERT$_{base}$, which only conducts classification loss on clean instances.
For every task of GLUE benchmark, every training is quite different such as \cite{adco} proposes a method to create adversarial examples on word-level 
and \cite{cline} uses external semantic knowledge to generate negative instances. process uses original checkpoint of BERT$_{base}$ with no train on other transfer tasks.
For tasks of MNLI, QQP, RTE, SST-2, accuracy is the judging criteria and F1 score is conducted to MRPC task.
COLA task uses evaluated by Matthews correlation coefficient (MCC).

Table 3 gives the result in development sets.
Obviously, compared with fine-tuned BERT$_{base}$, fine-tuning SCAL gives great improvement on every task.
For all tasks, SCAL leads to an improvement of 2.2\% on average compared to conventional fine-tuned BERT$_{base}$.
The most improvement is in the task of COLA, 5.4\% and the other two tasks are also improved greatly, 3.4\% for MRPC task and 2.2\% for RTE task.
We also conduct ablation experiment on these three tasks to demonstrate the combination of contrastive learning and adversarial training.
Table 4 shows the ablation result on three great improvement tasks.
Compared to model only used contrastive learning or adversarial training, SCAL gives improvement in MRPC and COLA tasks while a little lower in RTE task.

Short-text classification are evaluated in table 5.
The improvement on the two short text data sets has exceeded 1\% when compare to Bert$_{base}$.
And compared to Bert-ITPT-FiT \cite{finetunebert}, SCAL achieves better performance and shows state-of-the-art results.
The thing that SCAL model can work better in short-classification tasks means the adversarial examples as positive pairs is efficacious.

\subsection{Experiment Result on USCAL}

\textbf{Semantic textual similarity tasks} We evaluate on SentEval consisted of 7 tasks: STS 2012-2016, STS-B, SICK-Relatedness. 
The cosine similarity between sentence pairs in datasets is computed for calculating Spearman score.
Following the previous work, we calculate the all averaged Spearman scores for evaluation.
At the same time, we also use previous works to repeat the experiment, which confirms fair comparison.
The main result is shown in Table 6, using contrastive learning with adversarial training can significantly improve the performance in all datasets when compared 
with Bert$_{base}$, an improvement from 66.58\% to 77.29\%.
Although SimCSE gives a great improvement to 75.54\%, our model raises 1.7 \% from 75.54\% of SimCSE.
When the encoder changes to RoBERTa$_{base}$, our model also achieves improvement compared to other models.
The supervised training is not performed because we think transfer tasks shows unstable like SimCSE.

\subsection{Experiment Result of Robustness}
\textbf{Robustness} ANLI is a suitable dataset to evaluate the robustness model  as it is an adversarial training dataset.
We conduct the experiments with two forms as InfoBert: (i) training the model on two combined datasets (MNLI + SNLI), which means the model threated is unknown.
(ii) training models on both combined datasets and adversarial datasets (MNLI + SNLI + ANLI + FeverNLI), which in the case that the model threated is already known.

Result of the first form is summarized in Table 7. It is clear that the results of vanilla BERT and RoBERTa on adversarial dataset ANLI is poorly.
For instance, FreeLB BERT$_{large}$ only achieve 27.4\%, which is the lowest accuracy among all the models.
We also give the adversarial training results from \cite{infobert}, which uses FreeLB and InfoBERT to peform it.
In table 6, we just use the result of some methods in \cite{infobert}, where FreeLB achieves 27.4 \% using Bert and 30.8 \% using Roberta.
Whether using Bert or Roberta as the backbone, our model can improve by two percentage points compared to FreeLB.
For using the Bert as encoder, our model gets more improved accuracy in evaluation dataset and test dataset.
Specifically, our work achieves 32.4\% in ANLI test dataset while InfoBert gets 28.2\%.
And compared to InfoBert with RoBERTa, our model achieves average higher 36.3 \% in test dataset.

For the second setting, results are shown in Table 8, which indicates SCAL can further improve the performance for both Bert and RoBerta.
We training our models under the same batch-size 32 and the other model results are from the best result of \cite{infobert}.
Training with RoBERTa, FreeLB gets 56.2\% accuracy in the test dataset and the other two model, SMART and ALUM perform 57.1\% and 57.0\% accuracy.
Besides, InfoBert shows higher accuracy, both 58.3\% in the development dataset and test dataset, than the above models.
By adding SCAL, we can reach great robust accuracy of 58.6\% while evaluating in ANLI development dataset, outperforming the InfoBert model.
And the test accuracy is also close to the InfoBert model.

\begin{table*}[ht]
  \caption{Robust accuracy on the adversarial dataset ANLI. Models are trained on MNLI + SNLI datasets. 'A1-A3' means the three rounds of ANL and
  ANLI indicates the full of 'A1+A2+A3'. Where $\clubsuit $ indicates the result from \cite{infobert}}
  \centering
  \begin{tabular}{|c|c|c|c|c|c|c|c|c|c|}
    \hline
    
    \multirow{2}*{$Model$}& \multirow{2}*{$Method$} &\multicolumn{4}{c|}{$Dev$} &\multicolumn{4}{c|}{$Test$}\\
    \cline{3-10}
    \multicolumn{2}{|c|}{}&A1 & A2 &A3 &ANLI &A1 &A2 &A3 &ANLI \\
    \hline
    \multirow{2}*{$Bert$}  & FreeLB$\clubsuit $ &23.0 &29.0 & 32.2 & 28.3 &22.2 &28.5 &30.8 &27.4\\
    &InfoBERT$\clubsuit $ &28.3 &30.2 &33.8 &30.9 &25.9 &28.1 &30.3 &28.2\\
    &SCAL &29.5	&32.4	&32.8	&\textbf{31.9} &31	&33.4	&32.9 &\textbf{32.4}\\
    \hline
    \multirow{2}*{$RoBERTa$}  & FreeLB$\clubsuit $ &50.4 &28.0 &28.5 &35.2 &48.1 &30.4 &26.3 &34.4\\
    &InfoBERT$\clubsuit$ &48.4 &29.3 &31.3 &36.0 &50.0 &30.6 &29.3 &36.2\\
    &SCAL &48.2	&30.3	&30.9	&\textbf{36.4} &46.3	&33.0	&29.4 &\textbf{36.3}\\
    \hline
  \end{tabular}
\end{table*}

\begin{table*}[h]
  \caption{Robust accuracy on the adversarial dataset ANLI. Models are trained on MNLI + SNLI + ANLI(train) + FeverNLI datasets. 'A1-A3' means the three rounds of ANL and
  ANLI indicates the full of 'A1+A2+A3'. $\clubsuit $ directs the result from \cite{infobert}}
  \centering
  \begin{tabular}{|c|c|c|c|c|c|c|c|c|c|}
    \hline
    
    \multirow{2}*{$Model$}& \multirow{2}*{$Method$} &\multicolumn{4}{c|}{$Dev$} &\multicolumn{4}{c|}{$Test$}\\
    \cline{3-10}
    \multicolumn{2}{|c|}{}&A1 & A2 &A3 &ANLI &A1 &A2 &A3 &ANLI \\
    \hline
    \multirow{2}*{$Bert$}  & FreeLB$\clubsuit $ &60.3 &47.1 &46.3 &50.9 &60.3 &46.8 &44.8 &50.2\\
    &ALUM$\clubsuit $  &62.0 &48.6 &48.1 &52.6 &61.3 &45.9 &44.3 &50.1\\
    &InfoBERT$\clubsuit $ &60.8 &48.7 &45.9 &51.4 &63.3 &48.7 &43.2 &51.2\\
    &SCAL &64.5	&48.5	&45.8	&\textbf{52.5} &64.0	&47.6	&44.3 &\textbf{51.9}\\
    \hline
    \multirow{2}*{$RoBERTa$} 
    &FreeLB$\clubsuit $ &75.2 &47.4 &45.3 &55.3 &73.3 &50.5 &46.8 &56.2\\
    &SMART$\clubsuit $ &74.5 &50.9 &47.6 &57.1 &72.4 &49.8 &50.3 &57.1\\
    &ALUM$\clubsuit $ &73.3 &53.4 &48.2 &57.7 &72.3 &52.1 &48.4 &57.0\\
    &InfoBERT$\clubsuit $ &76.4 &51.7 &48.6 &58.3 &75.5 &51.4 &49.8 &\textbf{58.3}\\
    &SCAL &75.5	&50.3	&50.0	&\textbf{58.6} &71.1	&52.9	&50.3 &\textbf{58.1}\\
  
    \hline
  \end{tabular}
\end{table*}

\section{Conclusion}
In this work, we propose two framework, supervised  contrastive adversarial learning (SCAL) and  unsupervised SCAL (USCAL), to use contrastive adversarial learning, which largely improve the performance in supervised 
tasks and unsupervised tasks. 
Specifically, the proposed frameworks perform different generations of adversarial attack in different tasks, and the novelty is also the attack to embedding space.
Adversarial training generates difficult examples to learn, which also gives contrastive learning much harder positive instances.
The difficulty of learning can significantly improve the performance of previous models.
Specially, our model achieve state-of-the-art robust results under multiple adversarial datasets on NLI tasks.

We believe that the contrastive adversarial learning prompt will catch great attention in NLP. 
It gives a new way to reconsider the learning tasks and data augmentation in contrastive learning, which may lead to more new works and robust models.

\bibliographystyle{ACM-Reference-Format}
\bibliography{myref}

\appendix

\end{document}